\documentclass[11pt,a4paper]{article}
\usepackage{times,latexsym}
\usepackage{url}
\usepackage[T1]{fontenc}

\usepackage[acceptedWithA]{tacl2018v2}

\usepackage{xspace,mfirstuc,tabulary}

\newif\iftaclinstructions
\taclinstructionsfalse 
\iftaclinstructions
\renewcommand{\confidential}{}
\renewcommand{\anonsubtext}{(No author info supplied here, for consistency with
TACL-submission anonymization requirements)}
\newcommand{\instr}
\fi

\iftaclpubformat 

\else

\fi

\title{Narrative Question Answering with Cutting-Edge Open-Domain QA Techniques: A Comprehensive Study}

\usepackage{url}  
\usepackage{graphicx} 
\usepackage{booktabs}
\usepackage{amsfonts}
\usepackage{multirow}
\usepackage{amsmath}
\DeclareMathOperator*{\argmax}{arg\,max}

\usepackage{esvect}

\usepackage{array}
\newcommand{\PreserveBackslash}[1]{\let\temp=\\#1\let\\=\temp}
\newcolumntype{C}[1]{>{\PreserveBackslash\centering}p{#1}}
\newcolumntype{R}[1]{>{\PreserveBackslash\raggedleft}p{#1}}
\newcolumntype{L}[1]{>{\PreserveBackslash\raggedright}p{#1}}

\usepackage{xcolor}
\usepackage{arydshln}
\usepackage[switch]{lineno}

\usepackage{dblfloatfix}
\usepackage{balance}

\usepackage{extarrows} 

\newcommand{\proposedAnswer}{\tilde{\mathbf{A}}}
\newcommand{\realAnswer}{\mathbf{A}}
\newcommand{\question}{\mathbf{Q}}
\newcommand{\contextSet}{\mathcal{C}}
\newcommand{\filteredContextSet}[1][\question]{\mathcal{C}_{#1}}
\newcommand{\passage}[1][i]{\mathbf{C}_{#1}}

\author{Xiangyang Mou\Thanks{Equal contribution. XM built the whole system, implemented the data preprocessing pipeline, Hard EM ranker, and all the reader modules, and conducted all the QA experiments. CY implemented the unsupervised ICT ranker and the first working version of FiD, and was responsible for the final ranker module. MY is the corresponding author, who proposed and led this project, built the ranker code base (until the DS ranker), 
designed the question schema and conducted its related experiments and analysis in Part II.} \qquad Chenghao Yang$^{*}$
\qquad Mo Yu$^{*}$ \\
\medskip
\bf Bingsheng Yao \qquad \bf Xiaoxiao Guo \qquad \bf Saloni Potdar \qquad \bf Hui Su\\
\medskip
Rensselaer Polytechnic Institute \& IBM   \\
{\small \texttt{moux4@rpi.edu}\quad \texttt{gflfof@gmail.com}}
}

\date{}

\begin{document}
\maketitle
\begin{abstract}
Recent advancements in open-domain question answering (ODQA), i.e., finding answers from large open-domain corpus like Wikipedia, have led to human-level performance on many datasets.
However, progress in QA over book stories (Book QA) lags behind despite its similar task formulation to ODQA.
This work provides a comprehensive and quantitative analysis about the difficulty of Book QA: (1) We benchmark the research on the NarrativeQA dataset with extensive experiments with cutting-edge ODQA techniques. This quantifies the challenges Book QA poses, as well as advances the published state-of-the-art with a $\sim$7\% absolute improvement on Rouge-L. (2) We further analyze the detailed challenges in Book QA through human studies.\footnote{\url{https://github.com/gorov/BookQA}.} Our findings indicate that the event-centric questions dominate this task, which exemplifies the inability of existing QA models to handle event-oriented scenarios. 
\end{abstract}


\section{Introduction}\label{sec:introduction}

\begin{figure}[t!]
    \fontsize{14}{10}\selectfont
    \centering
    \includegraphics[width=0.44\textwidth]{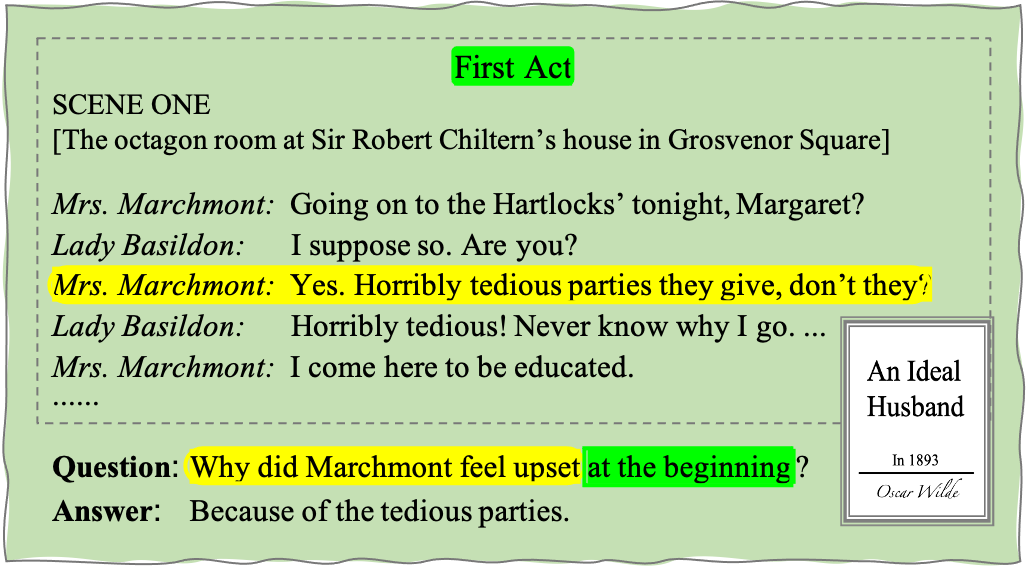}
    \vspace{-3mm}
    \caption{{An example of Book QA. The content is from the book \textit{An Ideal Husband}~\cite{wilde1916ideal}. The bottom contains a typical QA pair, and the highlighted text is the evidence for deriving the answer.}}
    \label{fig:bookqa_example}
    \vspace{-6mm}
\end{figure}

Recent Question-Answering (QA) models have achieved or even surpassed human performance on many challenging tasks, including single-passage QA\footnote{The SQuAD leaderboard~\cite{Rajpurkar_2018squad}: \url{rajpurkar.github.io/SQuAD-explorer}.} and open-domain QA (ODQA)\footnote{\citet{wang2020cluster,iyer2020reconsider}'s results on Quasar-T~\cite{dhingra2017quasar} and SearchQA~\cite{dunn2017searchqa}.}. Nevertheless, understanding rich context beyond text pattern matching remains unsolved, especially answering questions on narrative elements via reading books. One example is NarrativeQA~\cite{kovcisky2018narrativeqa} (Fig.~\ref{fig:bookqa_example}). Since its first release in $2017$, there has been no significant improvement over the primitive baselines. In this paper, we study this challenging Book QA task and shed light on the inherent difficulties.

Despite its similarity to standard ODQA tasks\footnote{Historically, open-domain QA meant ``QA on any domain/topic''. More recently, the term has been restricted to ``retrieval on a large pile of corpus''~\cite{chen2017reading}, so ``open-retrieval QA'' seems a better term here. However, to follow the recent terminology in the QA community, we still use ``open-domain QA'' throughout this paper.}, i.e., both requiring finding evidence paragraphs for inferring answers, the Book QA has certain unique \emph{challenges} \cite{kovcisky2018narrativeqa}:
\emph{(1)} the narrative writing style of book stories differs from the formal texts in Wikipedia and news, which demands a deeper understanding capability. The flexible writing styles from different genres and authors make the challenge severe;
\emph{(2)} the passages that depict related book plots and characters share more semantic similarities than the Wikipedia articles, which increases confusion in finding the correct evidence to answer a question; 
\emph{(3)} the free-form nature of the answers necessitates the summarization  ability from the narrative plots;  \emph{(4)} the free-form answers make it hard to obtain fine-grained supervision at passage or span levels;
and finally \emph{(5)} different paragraphs usually have logical relations among them.\footnote{We consider Challenge (5) more like an opportunity than challenges, and leave its investigation to future work.}

To quantify the aforementioned challenges,
we conduct a two-fold analysis to examine the gaps between Book QA and the standard ODQA tasks.
First, we benchmark the Book QA performance on the NarrativeQA dataset, with methods created or adapted based on the ideas of state-of-the-art ODQA methods~\cite{wang2018r,lin2018denoising,Lee_2019orqa,min2019discrete,guu2020realm,karpukhin2020dense}. 
We build a state-of-the-art Book QA system with a \emph{retrieve-and-read} framework, which consists of a \textbf{ranker} for retrieving evidence and a \textbf{reader} (i.e., QA model) to predict answers given evidence.
For the ranker model, we investigate different weakly supervised or unsupervised methods for model training with the lack of passage-level supervision.
For the reader model, we fill up the missing study and comparison among pre-trained generative models for Book QA, such as GPT-2~\cite{radford2019language} and BART~\cite{lewis2019bart}.
Then we investigate approaches to adapt to the book writing styles and to make use of more evidence paragraphs.
As a result, our study gives a $\sim$7\% absolute Rouge-L improvement over the published state-of-the-art.

Second, we conduct human studies to quantify the challenges in Book QA.
To this end, we design a new question categorization schema based on the types of reading comprehension or reasoning skills required to provide the correct answers. Precisely, we first define the basic \textbf{semantic units}, such as entities, event structures in the questions and answers. The question category thus \emph{determines the types of units and the relations between the units}.
We annotate $1,000$ questions accordingly and discover the significantly distinctive statistics of the NarrativeQA dataset from the other QA datasets, mainly regarding the focus of event arguments and relations between events. 
We further give performance decomposition of our system over the question categories, to show the detailed types of challenges in a quantitative way.

In summary, our comprehensive study not only improves the state-of-the-art with careful utilization of recent ODQA advancements, but also reveals the unique challenges in Book QA with quantitative measurements. 


\section{Related Work}

\paragraph{Open-Domain QA}
ODQA aims at answering questions from large open-domain corpora (e.g., Wikipedia). The recent work naturally adopts a ranker-reader framework~\cite{chen2017reading}. Recent success in this field mainly comes from improvement in the following directions:
(1) distantly supervised training of neural ranker models~\cite{wang2018r,lin2018denoising,min2019discrete,cheng2020probabilistic} to select relevant evidence passages for a question; (2) fine-tuning and improving the pre-trained LMs, like ELMo~\cite{elmo}, BERT~\cite{devlin2018bert}, as the rankers and readers; (3) unsupervised adaptation of pre-trained LMs to the target QA tasks~\cite{Lee_2019orqa,sun2019improving,xiong2019pretrained}.

\vspace{-1mm}
\paragraph{Book QA}
\label{sec:related_word_bookqa}
Previous works \cite{kovcisky2018narrativeqa,tay2019simple,frermann-2019-extractive} also adopt a ranker-reader pipeline. However, they have not fully investigated the state-of-the-art ODQA techniques. First, the NarrativeQA is a generative QA task by nature, yet the application of the latest pre-trained LMs for generation purposes, such as BART, is not well-studied. Second, lack of fine-grained supervision on evidence prevents earlier methods from training a neural ranking model, thus they only use simple BM25~\cite{robertson1995okapi} based retrievers. 
An exception is \cite{mou2020frustratingly} that constructs pseudo distance supervision signals for ranker training.
Another relevant work \cite{frermann-2019-extractive} uses book summaries as an additional resource to train rankers. However, this is different from the aim of Book QA task in answering questions solely from books, since in a general scenario the book summary cannot answer all questions about the book.
Our work is the first to investigate and compare improved training algorithms for rankers and readers in Book QA.


\section{Task Setup}

\subsection{Task Definition and Dataset}

Following~\citet{kovcisky2018narrativeqa}, we define the \textbf{Book QA} task as finding the answer $\realAnswer$ to a question $\question$ from a book, where each book contains a number of consecutive and logically-related paragraphs $\contextSet$. The size $|\contextSet|$ from different books varies from a few hundred to thousands.

All our experiments are conducted on the NarrativeQA dataset \cite{kovcisky2018narrativeqa}. It has a collection of 783 books and 789 movie scripts (we use books to refer to both of them), each containing an average of 62K words. Besides, each book has 30 question-answer pairs generated by human annotators in free-form natural language. Hence the exact answers are not guaranteed to appear in the books. NarrativeQA provides two different settings, the \textbf{summary} setting and the \textbf{full-story} setting. 
The former requires answering questions from book summaries from Wikipedia,
and the latter requires answering questions from the original books, assuming that the summaries do not exist.
Our Book QA task corresponds to the full-story setting, and we use both names interchangeably.

Following~\citet{kovcisky2018narrativeqa}, we tokenize the books with SpaCy\footnote{\url{https://spacy.io/}}, and split each book into non-overlapping trunks of 200 tokens.

\subsection{Baseline}
\label{ssec:baseline}
Following the formulation of the open-domain setting, we employ the dominating ranker-reader pipeline that first utilizes a \emph{ranker model} to select the most relevant passages $\filteredContextSet[\question]$ to $\question$ as evidence, 
\begin{equation} \label{eq:ranking}
    \small
    \filteredContextSet[\question] 
    = 
    \text{top-k}(\{P(\passage|\question) | \forall~ \passage \in \contextSet\});
\end{equation}
and then a \emph{reader model} to predict answer $\proposedAnswer$ given $\question$ and $\filteredContextSet[\question]$.

Our baseline QA systems consist of training different base reader models (detailed in Sec.~\ref{ssec:model_reader}) over the BM25 ranker.
We also compare with competitive public Book QA systems as baselines from \cite{kovcisky2018narrativeqa, frermann-2019-extractive, tay2019simple,frermann-2019-extractive,mou2020frustratingly} under the Narrative full-story setting, and a concurrent work~\cite{zemlyanskiy2021readtwice}.
As discussed in Section~\ref{sec:related_word_bookqa}, \citet{mou2020frustratingly} train a ranker with distant supervision (DS), i.e., the first analyzed ranker method (Fig. \ref{fig:analysis_cheatsheet}); \citet{frermann-2019-extractive} use exterior supervision from the book summaries, which is considered unavailable by design of the Book QA task.
Because the summaries are written by humans, the system can be viewed as benefiting from human comprehension of books.
Fig.~\ref{fig:model_cheatsheet} lists the details of our compared systems.

\begin{figure}[t!]
    \small
    \fontsize{6.5}{10.5}\selectfont
    \centering
    \setlength{\tabcolsep}{5pt}
    \begin{tabular}{
            C{0.23\textwidth}
            C{0.055\textwidth}
            C{0.07\textwidth}
            C{0.045\textwidth}}
        \toprule
        \multirow{2}{*}{\textbf{System}} & \textbf{trained ranker} & \textbf{pre-trained LM} & \textbf{extra data} \\
        \midrule
        IR+AttSum$^{\dagger}$ \cite{kovcisky2018narrativeqa}& & &\\
        IR+BiDAF$^{\ddagger}$ \cite{kovcisky2018narrativeqa}& & &\\
        IAL-CPG$^{\dagger}$ \cite{tay2019simple}&  & &    \\
        R$^3$$^{\ddagger}$ \cite{wang2018r}& $\pmb{\checkmark}$ & & \\
        BERT-heur$^{\ddagger}$ \cite{frermann-2019-extractive} &  $\pmb{\checkmark}$ & $\pmb{\checkmark}$ & $\pmb{\checkmark}$\\
        DS Ranker+GPT2$^{\dagger}$~\cite{mou2020frustratingly} &  $\pmb{\checkmark}$ & $\pmb{\checkmark}$ & \\
        DS Ranker+BERT$^{\ddagger}$~\cite{mou2020frustratingly} &  $\pmb{\checkmark}$ & $\pmb{\checkmark}$ & \\
        Our best QA system$^{\dagger}$ &  $\pmb{\checkmark}$ & $\pmb{\checkmark}$ & \\
        \bottomrule
    \end{tabular}
    \vspace*{-2mm}
    \caption{\small{Characteristics of the compared systems. $\dagger$/$\ddagger$ refers to generative/extractive QA systems. In addition to the standard techniques, \citet{wang2018r} use reinforcement learning to train the ranker; \citet{tay2019simple} use curriculum to train the reader.}}
    \vspace*{-3mm}
    \label{fig:model_cheatsheet}
\end{figure}

\subsection{Metrics}
Following previous works~\cite{kovcisky2018narrativeqa,tay2019simple,frermann-2019-extractive}, we use Rouge-L~\cite{lin-2004-rouge} as the main metric for both evidence retrieval and question answering.\footnote{For fair comparison, we lowercase the answers and remove the punctuation, and use the open-source nlg-eval library~\cite{sharma2017nlgeval}.} 
For completeness, Appendix \ref{app:append_full_results} provides results with other metrics used in the previous works, including Bleu-1/4~\cite{papineni2002bleu}, Meteor~\cite{banerjee2005meteor}, and the Exact Match (EM) and F1 scores that are commonly used in extractive QA.


\begin{figure*}[ht!]
    \small
    \fontsize{6.5}{10.5}\selectfont
    \centering
    \begin{tabular}{C{0.11\textwidth}
            C{0.17\textwidth}
            C{0.35\textwidth}
            C{0.29\textwidth}}
        \toprule
        \textbf{Target Module} &\textbf{Approach} & \textbf{Original Idea in ODQA} & \textbf{Our Improved Version for Book QA}   \\
        \midrule
        \multirow{4}{*}{Reader} &  \multirow{2}{*}{Book Prereading}&  \citet{sun2019improving} adapt GPT on general QA datasets;  \citet{xiong2019pretrained} adapt BERT on Wikipedia as entity prediction. & We propose to adapt BART to the narrative style with the text-infilling objective. \\
        \cdashline{2-4}
        & \multirow{2}{*}{Fusion-in-Decoder}& Proposed by \cite{izacard2020leveraging} as a new type of ODQA reader. &  We improve the decoder with attention over all the encoder states to capture cross-passage interaction. \\
        \midrule
        \multirow{4}{*}{Ranker} & \multirow{1}{*}{Heuristic distant supervision} &  \multirow{1}{*}{N/A} & N/A$^*$ \\
        \cdashline{2-4}
        & \multirow{2}{*}{Unsupervised ICT} & Proposed by \cite{Lee_2019orqa} as siamese network for both BERT pre-training and dense retrieval. & We improve the method with our book-specific training data selection. \\
        \cdashline{2-4}
        &Hard EM & Proposed by \cite{min2019discrete} for reader training. & We adapt the method for ranker training. \\
        \bottomrule
    \end{tabular}
    \vspace*{-2mm}
    \caption{{Summary of our inspected approaches in Analysis Part I. *We directly apply the heuristics from~\cite{mou2020frustratingly} for Book QA.}}
    \vspace*{-3mm}
    \label{fig:analysis_cheatsheet}
\end{figure*}

\section{Analysis Part I: Experimental Study}
This section describes our efforts of applying or adapting the latest open-domain QA ideas to improve Book QA ranker/reader models.
Fig.~\ref{fig:analysis_cheatsheet} summarizes our inspected approaches.
The experimental results quantify the challenges in Book QA beyond open-domain QA.

\subsection{QA Reader}     \label{ssec:model_reader}

\paragraph{Base Reader Models}
We study the usage of different pre-trained LMs on Book QA, including BART~\cite{lewis2019bart}, GPT-2~\cite{radford2019language}, T5~\cite{raffel2019exploring} and BERT~\cite{devlin2018bert}.
The first three are \emph{generative} readers and can be directly trained with the free-form answers as supervision. 
Specifically, during training we treat $\question \oplus \text{[SEP]} \oplus \filteredContextSet[\question]$ as input to generate answer $\realAnswer$, where $\text{[SEP]}$ is the special separation token and $\oplus$ is the concatenation operator.

For the \emph{extractive} reader (BERT), we predict the most likely span in $\filteredContextSet[\question]$ given the concatenation of the question and the evidence $\question \oplus \text{[SEP]} \oplus \filteredContextSet[\question]$.
Due to the generative nature of Book QA, the true answer may not have an exact match in the context. Therefore, we follow~\citet{mou2020frustratingly} to find the span $\mathbf{S}$ that has the maximum Rouge-L score with the ground truth $\realAnswer$ as the weak label, subject to that $\realAnswer$ and $\mathbf{S}$ have the same length (i.e. $|\mathbf{S}|=|\realAnswer|$).

\paragraph{Method 1: Book Prereading}
Inspired by the literature on the unsupervised adaptation of pre-trained LMs~\cite{sun2019improving,xiong2019pretrained}, we let the reader ``\textbf{preread}'' the training books through an additional pre-training step prior to fine-tuning with QA task.
This technique helps to better adapt to the narrative writing styles.

Specifically, we extract random passages from all training books to build a passage pool.
For each training iteration, we mask random spans from each passage, following the setting in~\cite{lewis2019bart}. The start positions of spans are sampled from a uniform distribution without overlapping. The length of each span is drawn from a Poisson distribution with $\lambda\!=\!3$. Each span is then replaced by a single \texttt{[mask]} token regardless of the span length. We mask $15\%$ of the total tokens in each passage. During the prereading stage, we use the masked passage as the encoder input and the raw passage as the decoder output to restore the raw passage in the auto-regressive way.

\paragraph{Method 2: Fusion-in-Decoder}
Recently~\citet{izacard2020leveraging} scale BART reader up to large number of input paragraphs. The method, \textbf{Fusion-in-Decoder (FiD)}, first concatenates each paragraph to the question to obtain a question-aware encoded vector, then merges these vectors from all paragraphs and feeds them to a decoder for answer prediction. 
FiD reduces the memory and time costs for encoding the concatenation of all paragraphs, and improves on multiple ODQA datasets.
FiD is an interesting alternative for Book QA, since it can be viewed as an integration of the ranker and reader, with the ranker absorbed in the separated paragraph encoding step. 

FiD trades cross-paragraph interactions for encoding more paragraphs. The single encoded vector per passage works well for extractive ODQA because the vector only needs to encode information of candidate answers. However, in Book QA, the answers may not be inferred from a single paragraph and integration of multiple paragraphs is necessary.
Therefore, in our approach, we concatenate the encoded vectors of all the paragraphs, and rely on the decoder's attention over these vectors to capture the cross-paragraph interactions.

\subsection{Passage Ranker}     \label{ssec:model_ranker}

\paragraph{Base Ranker Model}
Our ranker is a BERT-based binary classifier fine-tuned for evidence retrieval. It estimates the likelihood of each passage to be supporting evidence given a question $\question$.

Training the ranker models is difficult without high-quality supervision.
To deal with this problem, we investigate three approaches for creating pseudo labels, including \emph{distant supervision, unsupervised ranker training and Hard EM training}.

\paragraph{Method 1: Distant Supervision (DS)}
This is the baseline approach from~\cite{mou2020frustratingly}. It constructs DS signals for rankers in two steps: First, for each question $\question$, two BM25 rankers are used to retrieve passages, one with $\question$ as query and the other with both $\question$ and the true answer $\realAnswer$. Denoting the corresponding retrieval results as $\filteredContextSet[\question]$\footnote{For simplicity, we use the notation $\filteredContextSet[\question]$ here.} and $\filteredContextSet[\question + \realAnswer]$, the method samples the positive samples $\filteredContextSet[\question]^+$ from $\filteredContextSet[\question] \cap \filteredContextSet[\question + \realAnswer]$ and the negative samples $\filteredContextSet[\question]^-$ from the rest, with the ratio $\sigma \equiv |\filteredContextSet[\question]^+| / |\filteredContextSet[\question]^-|$ for each question $\question$ as a hyperparameter.

Second, to enlarge the margin between the positive and negative samples, the method applies a \textbf{Rouge-L filter} upon the previous sampling results to get the refined samples, $\mathcal{C}_\mathbf{Q}^{++}$ and $\mathcal{C}_\mathbf{Q}^{--}$:
\par 
\vspace*{-5mm}
{\small
    \begin{align}
        \filteredContextSet[\question]^{++} = 
        \left\{\max_{\mathbf{S}\subset \passage, |\mathbf{S}|=|\realAnswer|} {Sim}(\mathbf{S}, \realAnswer)>\alpha, \passage \in \filteredContextSet[\question]^+\right\}\,\,\,\nonumber
        \\
        \filteredContextSet[\question]^{--} = 
        \left\{\max_{\mathbf{S}\subset \passage, |\mathbf{S}|=|\realAnswer|} {Sim}(\mathbf{S}, \realAnswer)<\beta, \passage \in \filteredContextSet[\question]^-\right\}.\nonumber
    \end{align}
}%
$\mathbf{S}$ is a span in $\passage$, $Sim(\cdot,\cdot)$ is Rouge-L between two sequences. $\alpha$ and $\beta$ are hyperparameters.

\paragraph{Method 2: Unsupervised ICT Training}
Inspired by the effectiveness of Inverse Cloze Task (ICT) \cite{Lee_2019orqa} as an unsupervised ranker training objective, we use it to pre-train our ranker. The rationale is that we construct ``pseudo-question'' $q$ and ``pseudo-evidence'' $b$ from the same original passage $p$ and aim at maximizing the probability $P_{\mathrm{ICT}}(b | q)$ of retrieving $b$ given $q$, which is estimated using negative sampling as:
\begin{equation}
    \small
    P_{\mathrm{ICT}}(b | q)
    =
    \frac{\exp \left(S_{r e t r}(b, q)\right)}{\sum_{b^{\prime} \in B} \exp \left(S_{r e t r}\left(b^{\prime}, q\right)\right)}.
\end{equation}
$S_{retr}(\cdot,q)$ is the relevance score between a paragraph and the ``pseudo-question'' $q$. $b'$$\neq$$b$ is sampled from original passages other than $p$. 

The selection of ``pseudo-questions'' is critical to ICT training. To select representative questions, we investigate several filtering methods, and finally develop a book-specific filter \footnote{A unique filter is built for each book.}. Our method selects the top-scored sentence in a passage as a ``pseudo-question'' in terms of its total of token-wise mutual information against the corresponding book. The details can be found in Appendix~\ref{app:detail_ORQA}.

\paragraph{Method 3: Hard EM}
Hard EM is an iterative learning scheme. It was first introduced to ODQA by \citet{min2019discrete}, to find correct answer spans that maximize the reader performance.
Here we adapt the algorithm to ranker training.
Specifically, the hard EM can be achieved in two steps. At step $t$, the E-step first trains the reader with the current top-$k$ selections ${\filteredContextSet[Q]}^t$ as input to update its parameters $\Phi^{t+1}$; then derives the new positive passages ${\filteredContextSet[Q]^{+}}^{t+1}$ that maximizes the reader $\Phi^{t+1}$'s probability of predicting $\realAnswer$ (Eq.~\ref{eq:hardem_e2}).
The M-step updates the ranker parameter $\Theta$ (Eq.~\ref{eq:hardem_m}):

\par 
\vspace*{-5mm}
{\small
    \begin{align} 
        \label{eq:hardem_e2}
        {\filteredContextSet[Q]^{+}}^{t+1} &= k\text{-}\max_{\passage \in \contextSet} P(\realAnswer | \passage, \Phi^{t+1})   \\
        \label{eq:hardem_m}
        \Theta^{t+1} &= \argmax_{\Theta} P({\filteredContextSet[Q]^+}^{t+1} | \Theta^t).
    \end{align}
}%
In practice, \citet{min2019discrete} find that initialized with standard maximum likelihood training, the Hard EM usually converges in 1-2 EM iterations.



\section{Evaluation Part I: QA System Ablation}
We evaluate the overall Book QA system, and the individual modules on NarrativeQA. 

\medskip
\noindent\textbf{Implementation Details:}
For rankers, we initialize with \emph{bert-base-uncased}.
For readers, we use \emph{bert-base-uncased}, \emph{gpt2-medium}, \emph{bart-large} and \emph{T5-base}. The readers use top-3 retrieved passages as inputs, except for the FiD reader which uses top-10, making the readers have comparable time and space complexities.

\begin{table}[t!]
    \small
    \centering
    \setlength{\tabcolsep}{3pt} 
    \renewcommand{\arraystretch}{1} 
    \begin{tabular}{lcccccc} 
        \toprule
        \multirow{2}{*}{\bf System} & \multicolumn{2}{c}{\bf Rouge-L}\\
         & \bf dev & \bf test \\ 
        \midrule
        
        \multicolumn{3}{c}{\bf Public Extractive Baselines}    \\
        BiDAF~\cite{kovcisky2018narrativeqa}        & 6.33      & 6.22      \\
        R$^3$~\cite{wang2018r}                     & 11.40     & 11.90     \\
        DS-ranker + BERT~\cite{mou2020frustratingly} & 14.76 & 15.49  \\
        BERT-heur \cite{frermann-2019-extractive} & -- & 15.15     \\
        ReadTwice~\cite{zemlyanskiy2021readtwice} & 22.7 & 23.3 \\
        \midrule
        \multicolumn{3}{c}{\bf Public Generative Baselines}            \\
        Seq2Seq~\cite{kovcisky2018narrativeqa} & 13.29 & 13.15\\
        AttSum$^{*}$~\cite{kovcisky2018narrativeqa} & 14.86 & 14.02    \\
        IAL-CPG \cite{tay2019simple} & 17.33 & 17.67 \\
        DS-Ranker + GPT2 \cite{mou2020frustratingly} & 21.89 & 22.36 \\
        \midrule
        \multicolumn{3}{c}{\bf Our Book QA Systems}           \\
        BART-no-context (baseline) & 16.86 & 16.83 \\
        BM25 + BART reader (baseline)       & 23.16     & 24.47     \\
        Our best ranker + BART reader            & 25.83     & 26.95$^\dagger$     \\ 
        Our best ranker + our best reader            &  \bf 27.91   & \bf 29.21$^\dagger$     \\
        \quad \it repl ranker with oracle IR   & \it 37.75 & \it 39.32 \\
        \bottomrule
    \end{tabular}
    \vspace*{-2mm}
    \caption{{Overall QA
    performance (\%) in NarrativeQA Book QA setting. \textit{Oracle IR} combines question and true answers for BM25 retrieval. 
    We use an asterisk (*) to indicate the best results reported in \cite{kovcisky2018narrativeqa} with multiple hyper-parameters on dev set. The dagger ($^\dagger$) indicates significance with p-value < 0.01.}}
    \vspace*{-3mm}
    \label{tab:reader_model_performance}
\end{table}

\subsection{Overall Performance of Book QA}
We first show the positions of our whole systems on the NarrativeQA Book QA task. Table~\ref{tab:reader_model_performance} lists our results along with the state-of-the-art results reported in prior works (see Section \ref{ssec:baseline} and Fig.~\ref{fig:model_cheatsheet} for reference).
Empirically, our best ranker is from the combination of heuristic distant supervision and the unsupervised ICT training; our best reader is from the combination of the FiD model plus book prereading (with the top-10 ranked paragraphs as inputs). 
It is observed that specifically designed pre-training techniques play the most important role.
Details of the best ranker and reader can be found in the ablation study.

Overall, we significantly raise the bar on NarrativeQA by 4.7\% over our best baseline and 6.8\% over the best published one.\footnote{Appendix \ref{app:append_full_results} reports the full results, where we achieve the best performance across all of the metrics.} But there is still massive room for future improvement, compared to the upperbound with oracle ranker.
Our baseline is better than all published results with simple BM25 retrieval, showing the importance of reader investigation.
Our best ranker (see Section~\ref{ssec:exp_ranker} for details) contributes to $2.5$\% of our improvement over the baseline. 
Our best reader (see Section~\ref{ssec:exp_reader} for details) brings an additional $>$2\% improvement compared to the BART reader. 

We conduct a significance test for the results of our best system. There is no agreement on the best practice of the tests for natural language generation~\cite{clark2011better,dodge2019show}.  We choose the non-parametric bootstrap test, because it is a more general approach and does not assume specific distributions over the samples. 
For bootstrapping, we sample 10K subsets, the size of each is 1K.
The small p-value (< 0.01) shows the effectiveness of our best model.

As a final note, even the results with oracle IR are far from perfect. It indicates the limitation of text-matching-based IR; and further confirms the challenge of evidence retrieval in Book QA.

\subsection{Ranker Ablation}
\label{ssec:exp_ranker}
To dive deeper into the effects of our ranker training techniques in Sec.~\ref{ssec:model_ranker}, we study the intermediate retrieval results and measure their coverage of the answers. The coverage is estimated on the top-5 selections of a ranker from the baseline BM25's top-32 outputs, by both the maximum Rouge-L score of all the overlapped subsequences of the same length as the answer in the retrieved passages; and a binary indicator of the appearance of the answer in the passages (EM). 
Table~\ref{tab:ir_model_performance} gives the ranker-only ablation. On one hand, our best ranker improves both metrics. It also significantly boosts the BART reader compared to the DS-ranker~\cite{mou2020frustratingly}, as shown in Appendix~\ref{app:append_full_results}.
On the other hand, on top of the DS ranker,  
none of the other techniques can further improve the two ranker metrics significantly.
The ICT unsupervised training brings significant improvement over BM25. When adding to the DS-ranker, it brings slight improvement and leads to our best results.
Hard EM~\cite{min2019discrete} does not lead to improvements. Our conjecture is that generative readers does not solely generate purely matching-oriented signals, thus introduces noise in matching-oriented ranker training. 

The limited improvement and the low absolute performance demonstrate the difficulty of retrieval in Book QA. The gap between our best performance and the upper-bound implies that there is a large potential to design a more advanced ranker.

Additionally, we show that how much useful information our best ranker can provide to our readers in the whole QA system. 
In our implementation, the BART and FiD readers use top-3 and top-10 paragraphs from the ranker respectively.
The top-3 paragraphs from our best ranker give the answer coverage of 22.12\% EM and 49.83\% Rouge-L; and the top-10 paragraphs give 27.15\% EM and 56.77\% Rouge-L.
In comparison, the BM25 baseline has 15.75\%/43.44\% for top-3 and 24.08\%/53.55\% for top-10.
Therefore, our best ranker efficiently eases the limited-passage bottleneck brought by the ranker and benefits BART reader much more, which is consistent with our observations in Table~\ref{tab:reader_ablation}, Section~\ref{ssec:exp_reader}.

\begin{table}[t!]
    \small
    \centering
    \setlength{\tabcolsep}{4pt} 
    \begin{tabular}{lcc}
        \toprule
        \bf IR Method & \bf EM & \bf Rouge-L         \\
        \midrule
        \multicolumn{3}{c}{\bf Baseline Rankers}    \\
        BM25                        & 18.99 & 47.48 \\
        BERT DS-ranker~\cite{mou2020frustratingly}              
                                    & 24.26 & 52.68 \\
        \quad - Rouge-L filtering   & 22.63 & 51.02 \\
        \quad Repl BERT w/ BiDAF    & 21.88 & 50.64 \\
        \quad Repl BERT w/ MatchLSTM & 21.97 & 50.39 \\
        
        \midrule
        \multicolumn{3}{c}{\bf Our Rankers}    \\
        BERT ICT-ranker             & 21.29 & 50.35 \\
        BERT DS-ranker              \\
        \quad + Hard EM             & 22.45 & 50.50 \\
        \quad + ICT pre-training$^*$  & \bf 24.83 & \bf 53.19 \\
        \midrule
        \multicolumn{3}{c}{{\bf Oracle Conditions} }    \\
        Upperbound (BM25 top-32) & 30.81 & 61.40\\
        Oracle (BM25 w/ Q+A) & 35.75 & 63.92\\
        \bottomrule
    \end{tabular}
    \vspace*{-2mm}
    \caption{\small{Ranker performance (top-5) on dev set. Asterisk (*) indicates our best ranker used in Table \ref{tab:reader_model_performance}.
    }}
    \vspace*{-3mm}
    \label{tab:ir_model_performance}
\end{table}

\subsection{Reader Ablation}
\label{ssec:exp_reader}

\begin{table}[t!]
    \small
    \centering
    \setlength{\tabcolsep}{5pt} 
    \renewcommand{\arraystretch}{1} 
    \begin{tabular}{lcccccc} 
        \toprule
        \multirow{2}{*}{\bf System} & \multicolumn{2}{c}{\bf Rouge-L}\\
         & \bf dev & \bf test \\ 
        \midrule
        BM25 + BART reader (baseline)       & 23.16     & 24.47     \\
        \quad + BART-FiD reader             & 25.95     & --        \\
        Our ranker + BART reader            & 25.83     & 26.95     \\
        \quad + BART-FiD reader             & 26.27     & --        \\
        \quad repl BART w/ GPT-2            & 22.22     & --    \\
        \quad repl BART w/ T5               & 20.57     & --    \\
        \quad + book preread                & 26.82     & --    \\
        \quad \quad + BART-FiD Reader$^*$   & \bf 27.91 & \bf 29.21 \\
        \quad + book preread (\emph{decoder-only})  & 26.51     & --\\ 
        \bottomrule
    \end{tabular}
    \vspace*{-2mm}
    \caption{\small{Ablation of our Reader Model. Asterisk (*) indicates our best reader used in Table \ref{tab:reader_model_performance}.}}
    \vspace*{-4mm}
    \label{tab:reader_ablation}
\end{table}

Table~\ref{tab:reader_ablation} shows how the different reader techniques in Section~\ref{ssec:model_reader} contribute to the QA performance.

First, switching the BART reader to FiD gives a large improvement when using the BM25 ranker (2.8\%), approaching the result of ``our ranker + BART''.
This agrees with our hypothesis in Section~\ref{ssec:model_reader} Analysis 2, that FiD takes the roles of both ranker and reader.
Second, although the above result shows that FiD's ranking ability does not add much to our best ranker, our cross-paragraph attention enhancement still improves FiD due to better retrieval results (0.5\% improvement over ``our ranker + BART'').
Third, among all the generative reader models, BART outperforms GPT-2 and T5 by a notable margin. 
Finally, the book prereading brings consistent improvements to both combinations; and the combination of our orthogonal reader improvements finally gives the best results. We also confirm that the prereading helps decoders mostly, as only training the decoder gives comparable results.

\begin{figure*}[h!]
    \fontsize{8}{10.5}\selectfont
    \centering
    \begin{tabular}{lcp{5.9cm}p{5.8cm}}
        \toprule
        \textbf{SU Type} & \textbf{Sub Type} & \textbf{Description} & \textbf{Example}         \\
        \midrule
        \multirow{7}{*}{Concept}    
        
        & \multirow{3}{*}{Entity}     
        & {Standard named entities like person, location and organization names. Book-specific character names and their co-references are also included.} 
        & \multirow{3}{5.6cm}{\textbf{Q}: What is the name of \underline{Mortimer Treginnis}' sister? \textbf{A}: \underline{Brenda}} \\ 
        
        & Common Noun- 
        & \multirow{2}{5.6cm}{Common nouns or noun phrases that are universally used across books and other literature} 
        & \multirow{2}{5.6cm}{\textbf{Q}: What was Rodgers exposed to while investigating? \textbf{A}: \underline{Radioactive gas}}      \\
        
        &Phrases&&  \\
        
        & \multirow{2}{*}{Book-Specific} 
        & \multicolumn{1}{p{5.6cm}}{Common nouns or noun phrases that have special meanings or importance in the book of interests}  
        & {\textbf{Q}: Where do Anne and Philippa stay after their first year in college? \textbf{A}: \underline{Patty's place}}   \\
        
        \midrule
        
        \multirow{4}{*}{Event}      
        & \multirow{2}{*}{Event Expression}    
        & {Standard textual expression of event structures about "\emph{who did what to whom, when, where and how}"} 
        & \multirow{2}{6cm}{\textbf{Q}: In what way did \underline{Christopher atone for his sin}? \textbf{A}: \underline{He helped Will escape and accepted the punishment}} \\ 
        & \multirow{2}{*}{Event Name}  
        & \multicolumn{1}{p{5.6cm}}{Sometimes an important or famous event will be referred with a name} 
        & {\textbf{Q}: When did Harney and Charity kiss for the first time? \textbf{A}: On \underline{the trip to Nettleton}}    \\
        
        \midrule
        
        \multirow{8}{*}{Attribute}  
        & \multirow{2}{*}{States}     
        & \multicolumn{1}{p{5.6cm}}{The textual description of the state of an entity or concept as an attribute} 
        & {\textbf{Q}: Why does the princess agree to let Ermyntrude pretend to be her? \textbf{A}: Because she is \underline{timid}} \\ 
        
        & \multirow{2}{*}{Numerics} & \multirow{2}{5.6cm}{Standard attributes with numeric values}  
        & {\textbf{Q}: How may volumes has Darnley written on the origins of life? \textbf{A}: \underline{Three}} \\
        & \multirow{2}{*}{Descriptions} 
        & \multicolumn{1}{p{5.6cm}}{An attribute of an entity or concept which does not have a short attribute phrase to summarize} 
        & \multirow{2}{5.8cm}{\textbf{Q}: Why didn't Anne accept Gilberts proposal? \textbf{A}: \underline{She's dreaming of true love}} \\
        
        & \multirow{2}{*}{Book Attributes} 
        &         \multirow{2}{*}{Attributes of books themselves, like theme etc.}
        & {\textbf{Q}: Name the major theme used in the Adventures of Sherlock Holmes? \textbf{A}: \underline{Social injustice}} \\
        \bottomrule
    \end{tabular}
    \caption{\small{The definitions of semantic units (\textbf{SU}s). The \underline{underlined} texts represent the recognized SUs of the types.}}
    \label{fig:semantic_unit_definition}
\end{figure*}


\section{Analysis Part II: Human Study}
This section conducts in-depth analyses of the challenges in Book QA. We propose a \emph{new question categorization scheme} based on the types of comprehension or reasoning skills required for answering the questions; then conduct a human study on 1,000 questions. Consequently, the model performance per category provides further insights of the deficiency in current QA models.

\begin{figure*}[t!]
    \fontsize{8}{10.5}\selectfont
    \centering
    \begin{tabular}{lp{5.9cm}p{5.9cm}}
        \toprule
        \textbf{Question Type} & \textbf{Description} & \textbf{Example}         \\
        \midrule
        \multirow{2}{*}{Relation between Concepts}     
        &  {The question asks a relation a concept has, and expects another concept as the answer} 
        & {\textbf{Q}: What is the name of Mortimer Treginnis' sister? \textbf{A}: Brenda} \\
        
        \multirow{2}{*}{Attribute of Concept}       
        & {The question asks the value of an attribute a concept has, and expects an attribute value as the answer} 
        & \multirow{2}{5.9cm}{\textbf{Q}: How old is Conan?\\
        \textbf{A}: Around forty} \\
        
        \midrule
        
        \multirow{2}{*}{Event Argument - Concept}          
        & {The question asks an argument of an event, and expects the argument to be an concept} 
        &\multirow{2}{5.9cm}{\textbf{Q}: Where was Armitage discovered alive?\\	\textbf{A}: Italy} \\
        
        \multirow{2}{*}{Event Argument - Attribute}         
        & \multirow{2}{5.6cm}{Similar to the above, but asks for an argument that is an attribute}  
        & \multirow{2}{5.9cm}{\textbf{Q}: Where does Lady Dedlock believe Esther to be when the story starts? \textbf{A}: She believes her to be dead}
        \\ \\
        
        \multirow{2}{*}{Event Trigger}           
        & {A rare case whether the question asks the action, i.e., the trigger (like main verb) of an event}  
        & \multirow{2}{5.9cm}{\textbf{Q}: What does Conan do to the Pictish village? \\ \textbf{A}: He sets it on fire} \\
        \midrule
        
        \multirow{2}{*}{Causal Relation}         
        & {The question asks the cause of an event, and expects another event or a concept attribute as the answer} 
        & \multirow{2}{5.9cm}{\textbf{Q}: Why is Barabas angry at the Maltese governor? \textbf{A}: He robbed him} \\
        
        \multirow{2}{*}{Temporal Relation}       
        & {The question asks an event that has a type of temporal relations with another event or an attribute} 
        & \multirow{2}{5.9cm}{\textbf{Q}: What was Almayer doing when Mrs. Almayer snuck Nina away?	\textbf{A}: Drinking with the Dutch} \\
        
        \multirow{2}{*}{Nested Relation}         
        & {The question asks an argument of an event, while the argument's value is another event or an attribute} 
        & {\textbf{Q}: What did Dain vow to come back and help Almayer with? \textbf{A}: Finding the gold mine}\\
        \midrule
        
        \multirow{2}{*}{Book Attribute} 
        & \multirow{2}{5.6cm}{The question asks an attribute of the book itself} 
        & \multirow{2}{5.9cm}{\textbf{Q}: Where did the majority of the story occur ?\\ \textbf{A}: London} \\ \\
        \bottomrule
    \end{tabular}
    \caption{\small{The definitions of question types. Note that sometimes the answer repeats parts of the question like the last two examples in the second block, and we ignore these parts when recognizing the SUs in answers.}}
    \vspace*{-3mm}
    \label{fig:question_category_evidence_based}
\end{figure*}

\subsection{Question Categorization}

There have been many different question categorization schemes. Among them the most widely-used is intention-based, where an intention is defined by the \emph{WH}-word and its following word.
Some recent reasoning-focused datasets~\cite{yang2018hotpotqa,xiong2019tweetqa} categorize intents by the types of multi-hop reasoning or by the types of required external knowledge beyond texts.

However, all these previous schemes do not reasonably fit our analysis over narrative texts from two aspects:
(1) they only differentiate high-level reasoning types, which is useful in knowledge base QA (i.e., KB-QA) but fails to pinpoint the text-based evidence in Book QA; 
(2) they are usually entity-centric and overlook linguistic structures like events, while events play essential roles in narrative stories.
With this, we design a new systematic schema to categorize the questions in the NarrativeQA dataset.

\paragraph{Semantic Unit Definition}

We first identify a minimum set of basic semantic units, each describing one of the most fundamental components of a story. The set should be sufficient such that (1) each answer can be uniquely linked to one semantic unit, and (2) each question should contain at least one semantic unit. Our final set contains three main classes and nine subclasses (Fig.~\ref{fig:semantic_unit_definition}).

We merge the two commonly-used types in the previous analysis, named entities and noun phrases, into the \emph{Concept} class. The \emph{Event} class follows the definition in 
ACE 2005~\cite{walker2006ace}.
We also use a special sub-type ``\emph{Book Attribute}'' that represents the meta information or the global settings of the book, such as the era and the theme of the story in a book.

\paragraph{Question Type Definition}

On top of the semantic units' definition, each question can be categorized as a query that asks about either a semantic unit or a relation between two semantic units. We use the difference and split all the questions into nine types grouped in four collections (Fig.~\ref{fig:question_category_evidence_based}). 

\noindent $\bullet$ \textbf{Concept questions} that ask a \emph{Concept attribute} or a relation between two \emph{Concepts}. The most common types in most ODQA tasks (e.g., TriviaQA) and the QA tasks require multi-hop reasoning (e.g., ComplexQuestions and HotpotQA).

\noindent $\bullet$ \textbf{Event-argument questions} that ask parts of an event structure. This type is less common in the existing QA datasets, although some of them contain a small portion of questions in this class. The large ratio of these event-centric questions demonstrates the uniqueness of the NarrativeQA dataset.

\noindent $\bullet$ \textbf{Event-relation questions} that ask relations (e.g., causal or temporal relations) between two events or between an event and an attribute (a state or a description). This type is common in NarrativeQA, since events play essential roles in story narrations. A particular type in this group is the relation that one event serves as the argument of another event (e.g., \emph{how}-questions). It corresponds to the common linguistic phenomenon of (compositional) nested event structures.

\noindent $\bullet$ \textbf{Global-attribute questions} that ask \emph{Book Attribute}: As designed, it is also unique in Book QA.

\subsection{Annotation Details}

Five annotators are asked to label the semantic unit types and the question types on a total of 1,000 question-answer pairs.
There can be overlapped question categories for the same question. A major kind of overlaps is between the three event component types (trigger, argument - concept/attribute) and the three event relation types (causal, temporal and nested).
Therefore in the guideline, when the question can be answered with an event component, we ask the annotators to check if the question requires the understanding of event relations. If so, the question should be labeled with the event relation types as these are the more critical information for finding the answers.
Similarly, for the other rare cases of category overlaps, we ask the annotators to label the types that they believe are more important for finding the answers.

\begin{figure}[t!]
    \centering
    \includegraphics[width=0.45\textwidth]{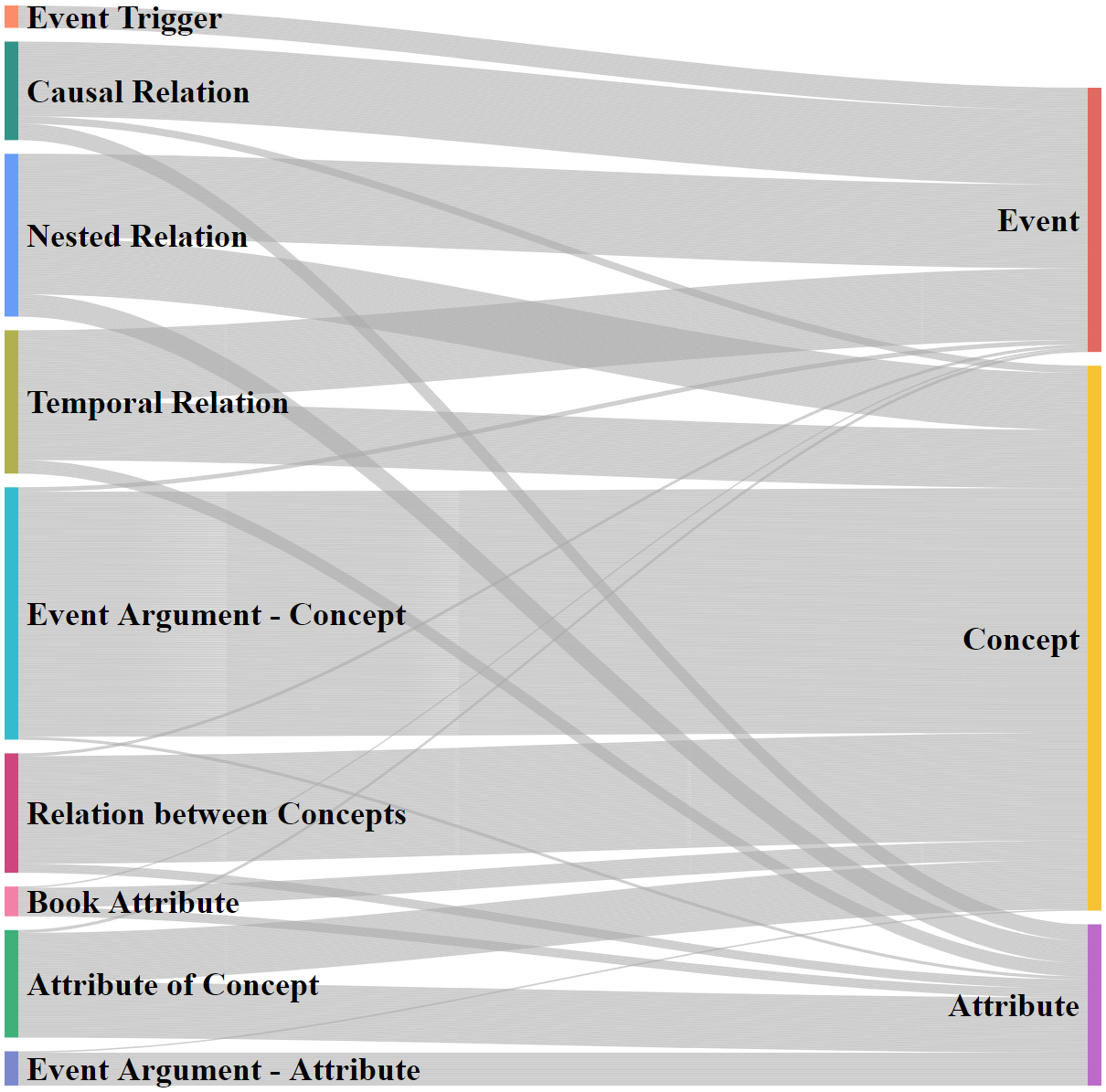}
    \vspace{-2mm}
    \caption{Visualization of the flow from the question types to their expected answer types.}
    \label{fig:annotation_category_flow}
    \vspace{-2mm}
\end{figure}

\paragraph{Correlation between question and answer types}
Figure~\ref{fig:annotation_category_flow} shows the ratios of answer types under each question type via a flow diagram.
Most question types correspond to a single major answer type, 
with a few exceptions:
(1) Most of the three event-relation questions have events as answers. A small portion of them have concepts or attributes as answers. This is either because the answers are state/description attributes; or because the answers are the arguments of one of the related events queried by the questions.
(2) The \emph{Relation b/w Concepts} type has some questions with attribute-typed answers. This is because the questions may ask the names of relations themselves, while some relation names are recognized as description-typed attributes.
(3) Most of \emph{Book Attribute} questions have concepts as answers, because they ask for the protagonists or the locations the stories occur at.

\paragraph{Annotation agreement}
A subset of 150 questions is used for quality checking, with each question labeled by two annotators. Table~\ref{tab:annotation_agreement} reports both the simple agreement rates and the Fleiss’ Kappa~\cite{fleiss1971measuring} $\kappa$s.
Our annotations reach a high agreement with around 90\% for question types and SU types and 80\% for SU sub-types, reflecting the rationality of our scheme.

\begin{table}[t!]
    \small
    \centering
    \begin{tabular}{lcc}
        \toprule
        \textbf{Category}   & \textbf{Simple Agreement(\%)} &\bf $\mathbf{\kappa}$(\%)  \\
        \midrule
        Question Type       & 88.0     & 89.9     \\
        SU Type             & 92.3     & 91.2     \\
        SU Sub Type         & 81.3     & 82.8     \\
        \bottomrule
    \end{tabular}
    \caption{\small{Annotation agreement. \textbf{SU}: Semantic Unit. ``SU Type'' and ``SU Sub Type'' are defined in Figure \ref{fig:semantic_unit_definition}. }}
    \vspace*{-1mm}
    \label{tab:annotation_agreement}
\end{table}

\subsection{Performance of Question Type Classification on the Annotated Data}
We conduct an additional experiment to study how well a machine learning model can learn to classify our question types based on question surface patterns.
We use the RoBERTa-base model that demonstrates superior on multiple sentence classification tasks.
Since our labeled data is small, we conduct a 10-fold cross validation on our labeled 1,000 instances. For each testing fold, we randomly select another fold as the development set and use the rest folds as training.

The final averaged testing accuracy is 70.2\%. Considering the inter agreement rate of 88.0\%, this is a reasonable performance, with several reasons for the gap:
(1) Our training data is too small and easy to overfit, evidenced by the performance gap between the training accuracy and development accuracy ($\sim$100\% versus 73.4\%). The accuracy can be potentially increased with more training data.
(2) Some of the ambiguous questions require the contexts to determine their types.
During labeling, our human annotators are allowed to read the answers for additional information, which leads to a higher upperbound performance.
(3) There is a small number of ambiguous cases, on which humans can use world knowledge while models are difficult to employ such knowledge. Therefore, the current accuracy can be potentially increased with a better model architecture.

\paragraph{Error Analysis and Lessons Learned}
Figure~\ref{fig:qtype_classify_confusion_matrix} gives major error types, which verifies the our discussed reasons above.
The majority of errors are the confusion between \emph{Event Argument - Concept} and \emph{Nested Relation}. The models are not accurate on the two types for several reasons: (1) Sometimes the similar question surface forms can take both concepts and events as an argument. In these cases, the answers are necessary for determining the question type. (2) According to our annotation guideline, we encourage the annotators to label event relations with higher priority, especially when the answer is a concept but serves as an argument of a clause. This increases the labeling error rate between the two types.
Another major error type is labeling \emph{Causal Relation} as \emph{Nest Relation}. This is mainly because some questions ask causal relations in an implicit way, on which human annotators have the commonsense to identify the causality but models do not.
The third major type is the failures in identifying the \emph{Attribute of Concept} and the \emph{Relation b/w Concepts} categories. As the attributes can be associated to some predicates, especially when they are descriptions, the models confuse them with relations or events.

The above observations provide insights on future refinement of our annotation guidelines, if people want to further enlarge the labeled data. For example, the \emph{Nested Relation} should be more clearly defined with comprehensive examples provided. In this way, the annotators can better distinguish them from the other types; and can better determine if the nested structure exists and whether to label the \emph{Event Argument} types.
Similarly, we could define clearer decision rules among relations, attributes and events, to help annotators distinguish \emph{Relation b/w Concepts}, \emph{Attribute of Concept} and \emph{Event Argument - Concept} types.

\begin{figure}[t!]
    \fontsize{6.5}{7.5}\selectfont
    \centering
    \begin{tabular}{llr}
        \toprule
        \textbf{Groundtruth Type}  & \textbf{Predicted Type} & \textbf{Freq}         \\
        \midrule
        \multirow{2}{*}{Relation between Concepts}    
        &  $\xlongrightarrow{\text{Fail}}    \text{Attribute of Concept}$           & 17/110     \\
        &   $\xlongrightarrow{\text{Fail}}    \text{Event Argument - Concept}$       & 12/110     \\
        
        \midrule
        
        \multirow{2}{*}{Attribute of Concept}    
        &  $\xlongrightarrow{\text{Fail}} \text{Relation between Concepts}$          & 21/120     \\
        &  $\xlongrightarrow{\text{Fail}} \text{Event Argument - Concept}$                   & 15/120     \\
        
        \midrule
        
        \multirow{3}{*}{Event Argument - Attribute}    
        &  $\xlongrightarrow{\text{Fail}} \text{Event Argument - Concept}$          & 6/34     \\
        &  $\xlongrightarrow{\text{Fail}}    \text{Attribute of Concept}$                & 6/34     \\
        &  $\xlongrightarrow{\text{Fail}}    \text{Temporal Relation}$              & 4/34     \\
        
        \midrule
        
        \multirow{1}{*}{Event Argument - Concept}    
        &  $\xlongrightarrow{\text{Fail}}    \text{Nested Relation} $               & 34/283     \\
        
        \midrule
        
        \multirow{2}{*}{Event Trigger}    
        &  $\xlongrightarrow{\text{Fail}}    \text{Nested Relation}$       & 4/18     \\
        &  $\xlongrightarrow{\text{Fail}}    \text{Event Argument - Concept}$       & 4/18     \\
        
        \midrule
        
        \multirow{1}{*}{Causal Relation}    
        &  $\xlongrightarrow{\text{Fail}}    \text{Nested Relation}$                & 17/126     \\
        
        \midrule
        
        \multirow{1}{*}{Nested Relation}    
        &  $\xlongrightarrow{\text{Fail}}    \text{Event Argument - Concept}$       & 35/154     \\

        \midrule
        
        \multirow{1}{*}{Book Attribute}    
        &  $\xlongrightarrow{\text{Fail}} \text{Attribute of Concept}$              & 3/29     \\

        \bottomrule
    \end{tabular}
    \caption{\small{Error analysis of question-type classification. We only list the major errors of each type (i.e., incorrect predicted types that lead to $>$10\% of the errors).}}
    \label{fig:qtype_classify_confusion_matrix}
\end{figure}

\begingroup
\setlength{\tabcolsep}{3pt}
\begin{table}[t!]
    \fontsize{8.7}{10}\selectfont
    \centering
    \begin{tabular}{lcccc}
        \toprule
        \multirow{2}{*}{\textbf{Question Type}} & \multirow{2}{*}{\textbf{Ratio(\%)}} & \multicolumn{2}{c}{\textbf{QA Rouge-L}}   &  \bf Ranker  \\
        & &\bf Gen &\bf Ext & \bf Rouge-L\\
        \midrule
        Relation b/w Concepts   & 11.0     & 40.48   & 24.46     & 63.76  \\
        Attribute of Concept    & 12.0     & 34.09   & 21.69     & 56.73  \\
        Event - Attribute       & 3.4      & 25.88   & 10.57     & 49.23  \\
        Event - Concept         & 28.3     & 27.35   & 15.73     & 62.15  \\
        Event - Trigger         & 1.8      & 29.63   & 9.28      & 37.56  \\
        Causal Relation         & 12.6     & 22.86   & 10.39     & 38.47  \\
        Temporal Relation       & 12.6     & 28.01   & 15.57     & 49.20  \\
        Nested Relation         & 15.4     & 23.02   & 8.44      & 48.93  \\
        Book Attribute          & 2.9      & 23.11   & 25.71     & 54.60  \\
        \bottomrule
    \end{tabular}
    \vspace*{-2mm}
    \caption{\small{Performance decomposition to question types of our best generative system (\textbf{Gen}, the best BART-based system), extractive system (\textbf{Ext}, the best BERT-based system, i.e., our best ranker + BERT reader), and ranker (BERT+ICT from Table~\ref{tab:ir_model_performance}).}}
    \label{tab:performance_question_category}
\end{table}
\endgroup

\begingroup
\setlength{\tabcolsep}{2.5pt}
\begin{table}[t!]
    \fontsize{8.5}{10}\selectfont
    \centering
    \begin{tabular}{lcccc}
        \toprule
        \multirow{2}{*}{\textbf{Answer Type}} & \multirow{2}{*}{\textbf{Ratio(\%)}} & \multicolumn{2}{c}{\textbf{QA Rouge-L}}   &\bf Ranker    \\
        & &\bf Gen &\bf Ext & \bf Rouge-L\\
        \midrule
        Concept - Entity            & 35.3   & 26.76   & 18.59      & 66.79  \\
        Concept - Common Noun       & 16.9   & 31.53   & 12.90      & 51.03  \\
        Concept - Book Specific     & 4.3    & 39.68   & 26.53      & 65.54  \\
        Event - Expression          & 25.1   & 24.62   & 11.50      & 39.40  \\
        Event - Name                & 2.8    & 24.79   & 5.54       & 42.88  \\
        Attribute - State           & 4.2    & 38.75   & 17.03      & 53.82  \\
        Attribute - Numeric         & 4.7    & 33.57   & 24.44      & 57.31  \\
        Attribute - Description     & 6.1    & 26.13   & 11.15      & 41.70  \\
        Attribute - Book Attribute  & 0.6    & 27.91   & 19.88      & 52.78  \\
        \bottomrule
    \end{tabular}
    \vspace*{-2mm}
    \caption{\small{Performance decomposition to answer types of our best generative/extractive systems and ranker. \textbf{Gen} and \textbf{Ext} are the same systems as in Table  \ref{tab:performance_question_category}.}}
    \label{tab:performance_answer_category}
\end{table}
\endgroup

\section{Evaluation Part II: QA System Performance Decomposition}
Table~\ref{tab:performance_question_category} presents both the ratio of each question type and our best generative and extractive performance on it. The ratios reflect NarrativeQA's unique focus on events, as $\sim$$75\%$ of the questions are relevant to the events in book stories. Specifically, $\sim$$34\%$ of the questions ask components of event structures (i.e., arguments or triggers) and $41\%$ ask relations between events (note that these questions may still require the  understanding of event structures). By comparison, the two dominating types in the other QA datasets, \emph{Concept Relation} and \emph{Concept Attribute}, only contribute to a ratio of $\sim$$23\%$. 
This agrees with human intuitions on the unique challenges in book understanding.

\medskip
\noindent\textbf{Most difficult question types:} The performance breakdown shows that all three event-relation types (\emph{Causal}, \emph{Temporal} and \emph{Nested}) are challenging to our QA systems.
The \emph{Causal} relation is the most difficult type with the lowest QA performance. The result confirms that the unique challenge in understanding event relations is still far from being well-handled by current machine comprehension techniques, even with powerful pre-trained LMs.
Moreover, these types can also be potentially improved by the idea of complementary evidence retrieval~\cite{wang2018evidence,iyer2020reconsider,mou2021complementary} in ODQA.

Besides the three event-relation types, the \emph{Event - Attribute} and \emph{Event - Triggers} are also challenging to the extractive system,
because the answers are usually long textual mentions of events or states that are not extractable from the passages.

\medskip
\noindent\textbf{Challenging types for reader:}
By checking the performance gaps of the generative system and the ranker, we can tell which types are difficult mainly for the reader.\footnote{Note that this analysis cannot confirm which types pose challenges to the ranker. This is because for event answers that are relatively longer and generative, there is a natural disadvantage on our pseudo ranker Rouge scores.} The \emph{Event - Concept} type poses more challenges to the reader, given that the ranker can perform well on them but the overall QA performance is low.
These questions are challenging mainly due to the current readers' difficulty in understanding the event structures, since their answers are usually extractable from texts.

\medskip
\noindent\textbf{Breakdown onto answer types:}
To better understand the challenges of non-extractable answers, we show the performance on each answer type in Table~\ref{tab:performance_answer_category}. 
The answers are mostly extractable when they are entities (including the book-specific terms and numeric values). On these types the extractive systems perform better and the two systems perform closer, compared to the other types.
In contrast, the answers are less likely to be extractable from the original passages when they are events, states, and descriptions. An interesting observation is that the \emph{Common Noun Phrases} type is also challenging for the extractive system.
It indicates that these answers may not appear in the texts with the exact forms, so commonsense knowledge is required to connect their different mentions.

\begin{table}[t!]
\setlength{\tabcolsep}{4pt}
    \small
    \centering
    \begin{tabular}{lcccccc} 
        \toprule
        \bf System & \multicolumn{2}{c}{\bf Full Data} & \multicolumn{2}{c}{\bf Event-Only}\\
        & \bf dev & \bf test & \bf dev & \bf test \\
        \midrule
        BERT+Hard EM & 58.1 &  58.8 & -- & -- \\
        Masque & -- & 54.7 & -- & -- \\
        BART Reader (ours) & \textbf{66.9} & \textbf{66.9} & 55.1 & 55.0 \\
        \bottomrule
    \end{tabular}
    \vspace*{-1mm}
    \caption{\small{Rouge-L scores under NarrativeQA summary setting. We list the best public extractive model BERT+Hard EM~\cite{min2019discrete} and the best generative model Masque~\cite{nishida2019multi} for reference. }
    }
    \vspace*{-3mm}
    \label{tab:reader_performance_on_summary}
\end{table}

\medskip
\noindent\textbf{Quantifying the challenge of event-typed answers to the reader:}
Table~\ref{tab:performance_answer_category} shows that the ranker performs poorly when the answers are events and descriptions.
This arouses a question -- whether the relatively lower QA performance is \emph{mainly due to the ranker's deficiency}; or \emph{due to the deficiency of both the ranker and the reader}. 

To answer this question, we conduct an experiment in the summary setting of NarrativeQA, to eliminate the effects of the ranker.
We create a subset of questions with event-typed answers if a question has either of its two answers containing a verb. 
This procedure results in a subset of 2,796 and 8,248 QA pairs in validation and test sets respectively.
We train a BART Reader with all training data in the summary setting, and test on both the full evaluation data and our event-only subsets.
Table~\ref{tab:reader_performance_on_summary} shows that the performance on the event-only subsets is about 12\% lower. 
The results confirm that questions with event-typed answers are challenging for both the reader and the ranker.


\section{Conclusion}

We conduct a comprehensive analysis on the Book QA task, taking the representative NarrativeQA dataset as an example.
Firstly, we design the Book QA techniques by borrowing the wisdom from the cutting-edge open-domain QA research and demonstrate through extensive experiments that (1) evidence retrieval in Book QA is difficult even with the state-of-the-art pre-trained LMs, due to the factors of rich writing style, recurrent book plots and characters, and the requirement of high-level story understanding; (2) our proposed approaches that adapt pre-trained LMs to books, especially the prereading technique for the reader training, are consistently helpful. 

Secondly, we perform a human study and find that (1) a majority of questions in Book QA requires understanding and differentiating events and their relations; (2) the existing pre-trained LMs are deficient in extracting the inter- and intra-structures of the events in the Book QA. Such facts lead us towards the event understanding task for future improvement over the Book QA task.


\section*{Acknowledgment}
This work is funded by RPI-CISL, a center in IBM’s AI Horizons Network, and the Rensselaer-IBM AI Research Collaboration (RPI-AIRC).

\bibliography{main-2849-Mou}
\bibliographystyle{acl_natbib}

\appendix

\section{Full Results on NarrativeQA}\label{app:append_full_results}


\begin{table*}[h!]
    \fontsize{8.9}{10}\selectfont
    \centering
    \setlength{\tabcolsep}{5pt} 
    \renewcommand{\arraystretch}{1} 
    \begin{tabular}{lcccccc} 
        \toprule
        \bf System & \bf Bleu-1 & \bf Bleu-4 & \bf Meteor & \bf Rouge-L & \bf EM & \bf F1 \\
        
        \midrule
        \multicolumn{7}{c}{\bf Public Extractive Baselines}    \\
        BiDAF~\cite{kovcisky2018narrativeqa}            & 5.82/5.68 & 0.22/0.25 &  3.84/3.72 & 6.33/6.22 & -- & --    \\
        R$^3$ \cite{wang2018r}                          &  16.40/15.70 &  0.50/0.49 & 3.52/3.47 & 11.40/11.90 & -- & -- \\
        BERT-heur~\cite{frermann-2019-extractive}       &  --/12.26 & --/2.06 &--/5.28 & --/15.15 & -- &--\\
        DS-Ranker + BERT~\cite{mou2020frustratingly}    & 14.60/14.46 &  1.81/1.38 & 5.09/5.03 & 14.76/15.49 & \bf 6.79/6.66 & 13.75/\textbf{14.45} \\
        ReadTwice(E)~\cite{zemlyanskiy2021readtwice}    & \bf 21.1/21.1  & \bf 3.6/4.0 & \bf 6.7/7.0 &   \bf 22.7/23.3  & --/-- & --/--  \\
        \midrule
        \multicolumn{7}{c}{\bf Our Extractive QA Models}\\
        BM25 + BERT Reader     & 13.27/13.84 & 0.94/1.07 & 4.29/4.59 & 12.59/13.81 & 4.67/5.26 & 11.57/12.55    \\
        \quad + HARD EM     & 14.39/-- & 1.72/-- & 4.61/-- & 14.10/--  & 5.92/-- & 12.92/--    \\
        \quad + ORQA        & {15.06}/14.25 & 1.58/1.30 & \textbf{5.28}/5.06 & \textbf{15.42}/15.22  & 6.25/6.19 & \textbf{14.58}/14.30    \\
        \quad + \it Oracle IR (BM25 w/ Q+A)&\it 23.81/24.01  & \it 3.54/4.01& \it 9.72/9.83&\it 28.33/28.72&\it 15.27/15.39&\it 28.42/28.55 \\
        \midrule
        \midrule
        & \multicolumn{3}{c}{\bf Public Generative Baselines}    \\
        AttSum (top-20)~\cite{kovcisky2018narrativeqa} & 19.79/19.06 & 1.79/2.11 & 4.60/4.37 & 14.86/14.02 & -- & --    \\
        IAL-CPG~\cite{tay2019simple} &23.31/22.92 & 2.70/2.47 & 5.68/5.59 & 17.33/17.67 & -- & --\\
        \quad - curriculum &20.75/-- & 1.52/-- & 4.65/--& 15.42/-- \\
        DS-Ranker + GPT2~\cite{mou2020frustratingly}   & 24.94/-- & 4.76/-- & 7.74/-- & 21.89/--  & 6.79/-- & 19.67/--    \\
        \midrule
        & \multicolumn{3}{c}{\bf Our Generative QA Models}               \\
        BM25 + BART Reader      & 24.52/25.30 & 4.28/4.65 & 8.68/9.25 & 23.16/24.47  & 6.28/6.73 & 21.16/22.28   \\
        \quad + DS-Ranker     & 24.91/25.22 & 4.28/4.60 & 8.63/8.82 & 23.39/24.10  & 6.67/6.93 & 21.31/21.93   \\
        \quad + HARD EM         & 25.83/-- & 4.48/-- & 8.75/-- & 24.31/--  & 7.29/-- & 21.91/--   \\ 
        \quad + Our Ranker            & 27.06/27.68 & 5.22/5.45 & 9.35/9.74 & 25.83/26.95  & 8.57/8.95 & 23.80/25.08   \\ 
        \quad\quad + Preread    & 28.54/-- & \textbf{6.13}/-- & 9.59/-- & 26.82/-- & 10.21/-- & 25.06/-- \\
        \quad\quad + FiD        & 28.04/-- & 5.66/-- & 9.49/-- & 26.27/--  & 9.20/-- & 24.29/--   \\
        \quad\quad + FiD + Preread & \textbf{29.56/29.98} & 6.11/\textbf{6.31} & \textbf{10.03/10.33} & \textbf{27.91/29.21} & \textbf{10.45/11.16} & \textbf{26.09/27.58}    \\
        \quad + \it Oracle IR (BM25 w/ Q+A)   & \textit{35.04/36.41} & \textit{8.84/9.08} & \textit{14.78/15.07} & \textit{37.75/39.32} & \textit{15.78/17.27} & \textit{37.71/38.73}     \\ 
        \hdashline[5pt/5pt]
        BM25 + GPT-2 Reader      & 24.54/-- & 4.74/-- & 7.32/-- & 20.25/--  & 5.12/-- & 17.72/--     \\
        \quad + Our Ranker        & 24.85/-- & 5.01/-- & 7.84/-- & 22.22/--  & 7.29/-- & 20.03/--    \\
        \quad + \it Oracle IR (BM25 w/ Q+A)   & \it 33.18/32.95 & \it 8.16/7.70 & \it 12.35/12.47 & \it 34.83/34.96 & \it 17.09/15.98 & \it 33.65/33.75\\ 
        \hdashline[5pt/5pt]
        BM25 + T5 Reader        & 19.28/-- & 3.67/-- & 6.62/-- & 16.89/--  & 4.17/-- & 15.47/--    \\
        \quad + Our Ranker            & 22.35/-- & 4.31/-- & 7.59/-- & 20.57/-- & 6.13/-- & 18.48/--     \\ 
        \quad + \it Oracle IR (BM25 w/ Q+A)   & \textit{31.06/31.49} & \textit{8.36/8.32} & \textit{12.61/12.93} & \textit{31.18/32.43} & \textit{12.77/12.84} & \textit{31.23/32.18}     \\ 
        \bottomrule
    \end{tabular}
    \vspace*{-2mm}
    \caption{{Full results on NarrativeQA dev/test set (\%) under the Book QA setting. We perform model selection based on the Rouge-L score on development set. DS is short for Distant Supervision in Sec.~\ref{ssec:model_ranker}.}}
    \vspace*{-2mm}
    \label{fig:full_experiments}
\end{table*}

Table~\ref{fig:full_experiments} gives full results with different metrics. 

\section{Details of ICT Training Data Creation}
\label{app:detail_ORQA}
Our pilot study shows that uniformly sampling the sentences and their source passages as ``pseudo-questions'' (PQs) and ``pseudo-evidences'' (PEs) does not work well.
Such selected PQs have high probability to be casual, e.g., ``\emph{Today is sunny}'', thus are not helpful for ranker training.

To select useful PQs, we define the following measure $f(s, b_j)$ to level the affinity between each candidate sentence $s$ and the book $b_j$:
\begin{equation}
    \small
    f(s, b_j) = \sum_{w_{ik} \in s}{\texttt{pmi}(w_{ik}, b_j)}
\end{equation}
where $\texttt{pmi}(w_k, b_j)$ is the word-level mutual-information between each word $w_{ik} \in s$ and the book $b_j$. Intuitively, $\texttt{pmi}(w_k, b_j)$ can be seen as the  ``predictiveness'' of the word $w_k$ with respect to the book $b_j$, and $f(s, b_j)$ measures the aggregated `importance'' for $s$. Consequently, the sentence $s$ with the highest $f(s, b_j)$ from each passage $p_n$ will be selected as the PQ; the corresponding $p_n$ with the PQ removed becomes the positive sample; whereas the corresponding negative samples from the same book $b_j$ will be the top-$500$ passages  (exclusive of the source passage $p_n$) with the highest TF-IDF similarity scores to the PQ. 

During sampling, we filter out stopwords and punctuation when computing $f(s, b_j)$. In movie scripts, the instructive sentences like ``\emph{SWITCH THE SCENARIO}'' that have poor connections to its source passages are also ignored. 
Finally, we require each PQ contain a minimum number of 3 non-stopwords.


\end{document}